\documentclass[conference]{IEEEtran}
\IEEEoverridecommandlockouts
\usepackage{cite}
\usepackage{amsmath,amssymb,amsfonts}
\usepackage{bm}
\usepackage{graphicx}
\usepackage{textcomp}
\usepackage{xcolor}
\usepackage{algorithm}
\usepackage{algcompatible}
\usepackage{algpseudocode}
\usepackage[english]{babel}
\usepackage{booktabs}
\usepackage{hyperref}
\usepackage{cleveref}[capitalise] 
\usepackage{float}
\newtheorem{theorem}{\textbf{Theorem}}

\usepackage{etoolbox}
\makeatletter
\patchcmd{\@makecaption}
  {\scshape}
  {}
  {}
  {}
\makeatletter
\patchcmd{\@makecaption}
  {\\}
  {:\ }
  {}
  {}
\makeatother
\usepackage[skip=1pt,font=small,labelfont=bf]{caption}

\newcommand{\bb}{\mathbf}  

\newcommand{\ourmethod}{V-PRISM}
\newcommand{\fullacronym}{\underline{V}olumetric, \underline{P}robabilistic, and \underline{R}obust \underline{I}nstance \underline{S}egmentation \underline{M}aps}

\algnewcommand\RETURN{\State \algorithmicreturn}%

\def\BibTeX{{\rm B\kern-.05em{\sc i\kern-.025em b}\kern-.08em
    T\kern-.1667em\lower.7ex\hbox{E}\kern-.125emX}}
\begin{document}

\title{\LARGE \bf 
    \ourmethod{}: Probabilistic Mapping of Unknown Tabletop Scenes
}

\author{Herbert Wright$^{1}$ \and Weiming Zhi$^{2}$ \and Matthew Johnson-Roberson$^{2}$ \and Tucker Hermans$^{1, 3}$
\thanks{$^{1}$ University of Utah Robotics Center and Kahlert School of Computing, University of Utah, Salt Lake City, UT, USA}%
\thanks{$^{2}$ Robotics Institute, Carnegie Mellon University, Pittsburgh, PA, USA}%
\thanks{$^{3}$ NVIDIA Corporation, Santa Clara, CA, USA}%
}

\frenchspacing
\maketitle

\begin{abstract}

The ability to construct concise scene representations from sensor input is central to the field of robotics. This paper addresses the problem of robustly creating a 3D representation of a tabletop scene from a segmented RGB-D image. These representations are then critical for a range of downstream manipulation tasks. Many previous attempts to tackle this problem do not capture accurate uncertainty, which is required to subsequently produce safe motion plans. In this paper, we cast the representation of 3D tabletop scenes as a multi-class classification problem. To tackle this, we introduce \ourmethod{}, a framework and method for robustly creating probabilistic 3D segmentation maps of tabletop scenes. Our maps contain both occupancy estimates, segmentation information, and principled uncertainty measures. We evaluate the robustness of our method in (1) procedurally generated scenes using open-source object datasets, and (2) real-world tabletop data collected from a depth camera. Our experiments show that our approach outperforms alternative continuous reconstruction approaches that do not explicitly reason about objects in a multi-class formulation.

\end{abstract}

\section{Introduction}

As robots continue to be deployed in the world, there is an ongoing need for methods that allow them to safely and robustly operate in unknown, noisy scenes. Many such scenes contain objects that robots must delicately move around or interact with to complete their assigned tasks. 
The planning techniques for such tasks often require an accurate 3D map of the objects within the scene. These are often unseen objects with unknown geometry that are only partially observed. 

The safe operation of robots necessitates not only accuracy but also introspection and uncertainty-awareness. These notions of uncertainty about the geometry of the scene can then be incorporated into downstream motion planning solvers for added robustness and safety. However, many learning algorithms typically used in robot learning, such as neural networks, lack the ability to reason about uncertainty and confidently predict incorrect labels \cite{goodfellow2014explaining, nguyen2015deep}. In this work, we take a Bayesian learning approach which captures uncertainty in a principled manner. 

\renewcommand{\thefootnote}{$\star$} 

\begin{figure}[t]
    \centering
    \includegraphics[width=\linewidth]{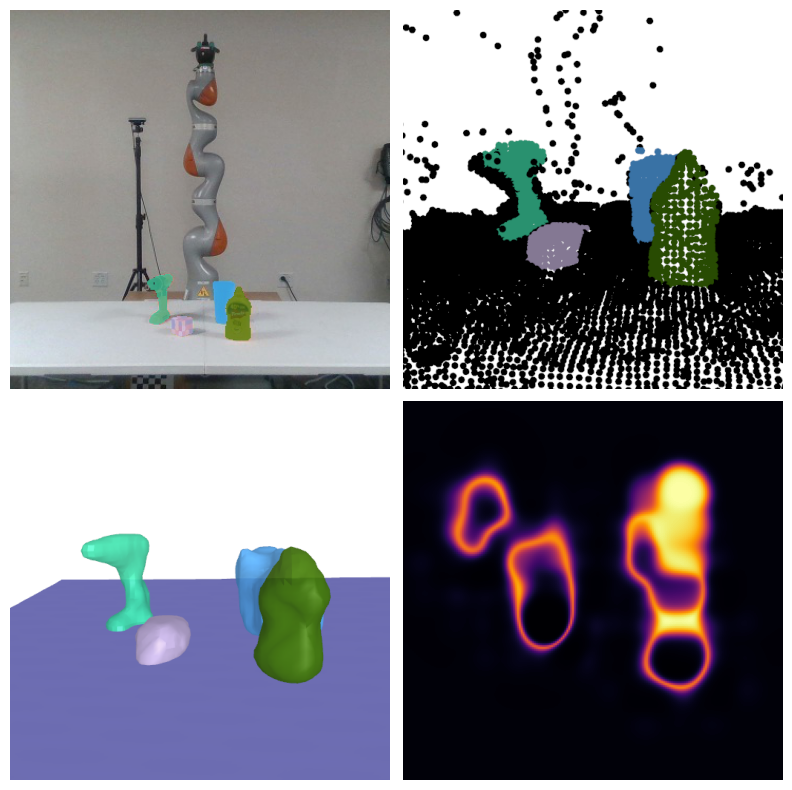}
    \caption{Our method takes a segmented ({\bf top left}) point cloud observation (\textbf{top right}) and builds a continuous probabilistic map. This map can be used to reconstruct the scene (\textbf{bottom left}) or measure uncertainty about the scene (\textbf{bottom right}). The heat map shows uncertainty in a 2D slice parallel with the table plane. Uncertainty is high in occluded areas.}
    \label{Fig:img1}
\end{figure}

We propose \ourmethod{}: \fullacronym{}\footnote{Code is available at \url{https://github.com/Herb-Wright/v-prism}}. \ourmethod{} is a framework for building differentiable segmentation and occupancy maps of tabletop scenes that contain multiple unseen objects. Importantly, our method results in maps with a principled and understandable uncertainty metric. To construct these maps, we rely on depth measurements with corresponding instance segmentations. Such instance segmentations can be easily obtained for real-world scenes using pre-existing models such as those proposed in \cite{kirillov2023segment, xie2021unseen, he2017mask}. We take inspiration from Bayesian Hilbert Maps (BHMs) \cite{senanayake2017bayesian} and transform points into an embedding induced by a set of chosen hinge points in order to perform Bayesian updates to our map. These updates are made in a variational manner with an expectation maximization (EM) algorithm. In order to effectively learn the geometry of the scene, we propose a negative sampling method for encoding depth sensor information in object-centric scenes. The learned map can be used to reconstruct the objects in the scene as well as measure the uncertainty about the geometry in different areas of the scene. This is pictured in \Cref{Fig:img1}.

We evaluate our method in simulation and the real world. The simulation scenes are constructed using objects from existing mesh datasets and placing them in a random configuration within a simulator similar to \cite{agnew2021amodal}. We run extensive experiments and report measurements of two commonly used metrics: intersection over union (IoU) and Chamfer distance of reconstructed meshes. Qualitative reconstructions and uncertainty estimates are computed on real world scenes of objects belonging to unknown classes to demonstrate robustness to noise associated with real-world cameras.

Concretely, our technical contributions include:

\begin{itemize}
    \item The formulation of 3D scene reconstruction as a multi-class mapping problem
    \item A principled \emph{Bayesian} framework to learn \emph{continuous} maps for tabletop scene representation
    \item An object-centric sampling method that enables accurate and efficient reconstructions 

\end{itemize}

Our paper is organized as follows. In \Cref{sec:related}, an overview of related research is provided. In \Cref{sec:preliminaries}, we review the basics of BHMs and formulate the problem our method aims to solve. Following this, \Cref{sec:overview} provides a high level overview of our proposed method. The math behind our EM algorithm and Bayesian model are discussed \Cref{sec:softmax}. We propose a novel negative sampling method in \Cref{sec:sampling} specifically for object-centric mapping. In \Cref{sec:experiments} We justify our decisions through quantitative and qualitative experiments. We also give qualitative examples of how our method provides desireable uncertainty measurements. This is followed by brief conclusion in \Cref{sec:conclusion}.

\section{Related Works}\label{sec:related}
{\bf 3D Mapping.} Constructing a 3D map of an environment has been a common problem in the field of robotic perception. Voxel based approaches such as Truncated Signed Distance Functions \cite{oleynikova2016signed} and OctoMaps \cite{hornung2013octomap} are a common approach to the problem. Hilbert Maps \cite{ramos2016hilbert}, and their extension to Bayesian Hilbert Maps \cite{senanayake2017bayesian} are both methods for mapping 3D environments. Recently, learning based approaches such as \cite{sucar2021imap} have grown in popularity. While these works serve as inspiration to our work, mapping a surrounding environment is a different problem than the one we approach in this paper.

{\bf Multiple View Synthesis.} One common way of reconstructing the geometry of an object or scene is to combine multiple camera views and observations into a 3D representation. One such technique is 3D-R2N2 \cite{choy20163d}, which creates a voxelized representation of objects using a recurrent neural network. More recently, Neural Radiance Fields \cite{mildenhall2021nerf} and variants such as Plenoxels \cite{fridovich2022plenoxels} learn an implicit density field using multiple views without depth information. Another learning based method for reconstructing scenes with multiple objects from multiple views is introduced in \cite{sucar2020nodeslam}, where a voxel encoder-decoder network is used. 3D Gaussian Splatting has also been used to reconstruct density fields from multiple images \cite{kerbl20233d}. While there has been a lot of interesting work around recovering 3D geometry by synthesizing multiple views, this work is directed at the harder problem of recovering 3D geometry using only a single view. 

{\bf Reconstructing Single Objects.} Many methods have been proposed to reconstruct single objects. In \cite{liu2022robust}, superquadrics are fit to the observed points to generate an objects geometry. Another method for recovering object geometry is Gaussian Process Implicit Surfaces \cite{dragiev2011gaussian} which implicitly reconstructs objects using a handful of surface points and their corresponding normals specifically for the robotic task of grasping. More recently, deep learning has been utilized to predict the geometry of objects from a single view. One deep learning approach is proposed in \cite{tulsiani2018factoring}, where a single image is used as input to a neural network that predicts object geometry. DeepSDF \cite{park2019deepsdf} and PointSDF \cite{van2020learning} are both learning based approaches that predict a signed distance function given a point cloud. Occupancy Networks \cite{mescheder2019occupancy} also take a single point cloud or image as input, but instead predict an occupancy function that maps each point in space to a probability of being occupied. Other work, such as \cite{rustler2023efficient} utilizes different modes of input like tactile measurements. The GenRe algorithm aims to predict the geometry of specifically unknown object classes as proposed in \cite{zhang2018learning}, where the authors explain that trying to generalize to unseen classes is much more difficult than traditional reconstruction. Occlusion is also a problem for many single-object reconstruction methods, as they generally assume that no other objects are present to partially obstruct an object.

{\bf Reconstructing Multiple Objects.} Some methods have been proposed that attempt to reconstruct scenes containing multiple objects with some occlusion. 3DP3 \cite{gothoskar20213dp3} is a method specifically for multi object scenes that assumes known object classes and uses probabilistic programming to reconstruct the scene. In \cite{li2019silhouette}, silhouettes are used to refine the voxelized predicted geometry of objects under occlusion. Another reconstruction technique proposed recently is ARM \cite{agnew2021amodal}, where scenes are encoded as voxels and a loss function is used that includes terms for connectivity and stability in order to increase generalizability. In contrast to our method, these approaches either assume known object classes or do not accurately measure the uncertainty over the scene.

\section{Preliminaries}\label{sec:preliminaries}

\subsection{Sigmoid Bayesian Hilbert Maps}\label{subsec:bhm}

{\bf Hilbert Maps.} Introduced in \cite{ramos2016hilbert}, Hilbert Maps are a method for continuous occupancy mapping of a robotic environment. A map $m: \mathbb R^d \rightarrow \mathbb [0, 1]$ is built from a feature transform $\phi(\bb x)$ and a set of $n'$ point observations $\{\bb x_i\}_{i \in [n']}$ from a sensor at position $\bb o$. The observed point cloud is labeled with $y_i = 1$ and unoccupied negative samples drawn on the line segments between each $\bb x_i$ and $\bb o$ are labeled with $y_i = 0$. The observed points, sampled points, and labels form the data $\{(\bb x_i, y_i)\}_{i \in [n]}$. Gradient descent is used to find the optimal weights for a map of the form
\begin{equation*}
    m(\bb x) = \sigma(\bb w^\top \phi(\bb x)) = (1 + \exp(-\bb w^\top \phi(\bb x)))^{-1},
\end{equation*}
where $\sigma: \mathbb{R}\rightarrow (0,1)$ is the sigmoid function. This is equivalent to performing logistic regression over the transformed points $\{(\phi(\bb x_i), y_i)\}_{i \in [n]}$.

Usually, the feature transform $\phi$ is constructed from a kernel function $k$ and a set of hinge points $\bb h_1, ..., \bb h_m \in \mathbb{R}^{3}$. Usually, these hinge points are chosen to be an evenly spaced 3D grid of points. The feature transform is then given by:
\begin{equation}\label{eq:hinge}
    \phi(x) = \begin{bmatrix} k(\bb x, \bb h_1) \\ k(\bb x, \bb h_2) \\ ... \\ k(\bb x, \bb h_m) \\ 1\end{bmatrix}.
\end{equation}


{\bf Bayesian Extension.} Hilbert Maps were extended to the Bayesian setting in \cite{senanayake2017bayesian}. Instead of an individual weight vector, the weight is treated as a normally distributed random variable,  $\bb w \sim P(\bb w)$. Variational Bayesian logistic regression as described in \cite{jaakkola1997variational} is then performed over data $D = \{(\phi(\bb x_i), y_i)\}_{i \in [n]}$ in order to obtain the approximate posterior distribution:
\begin{equation*}
    \hat{P}(\bb w | D) \propto Q(D | \bb w; \xi) P(\bb w) \approx P(D | \bb w) P(\bb w),
\end{equation*}
where the variational parameter $\xi$ is introduced. The method relies on an EM algorithm that alternates between calculating the posterior $\hat P(\bb w | D) = \mathcal N (\hat{\mu}, \hat{\Sigma})$ from an approximate likelihood function and obtaining a better likelihood approximation. The specific approximation used for the likelihood takes the form of a normal distribution, and ensures that the approximated likelihood is conjugate to a normal prior $P(\bb w) = \mathcal N (\bar{\mu}, \bar\Sigma)$.

Once the posterior weight distribution is obtained, the map $m$ is defined by the expectation:
\begin{equation*}
    m(\bb x) = \mathbb E_{\bb w} [\sigma(\bb w^\top \phi(\bb x))].
\end{equation*}
Because there is not an analytic solution for this expectation, approximations are used. The most common approximation is
\begin{equation}\label{eq:sigexpectation}
    \mathbb E_{\bb w} [\sigma(\bb w^\top \phi(\bb x))] \approx \sigma \left(\frac{\mathbb E_{\bb w}[\bb w^\top \phi(\bb x)]}{\sqrt{1 + \frac{\pi}{8} \text{Var}(\bb w^\top \phi(\bb x))}}\right),
\end{equation}
which is easily obtained for any $\bb w$ following a normal distribution.

Extensions of BHMs include Bayesian treatment of kernel parameters and hinge point placement \cite{senanayake2018automorphing}, fusing two BHMs \cite{zhi2019continuous}, and mapping environments with moving actors \cite{senanayake2018building}.




\subsection{Problem Formulation}

\begin{figure}
    \centering
    \includegraphics[width=\linewidth]{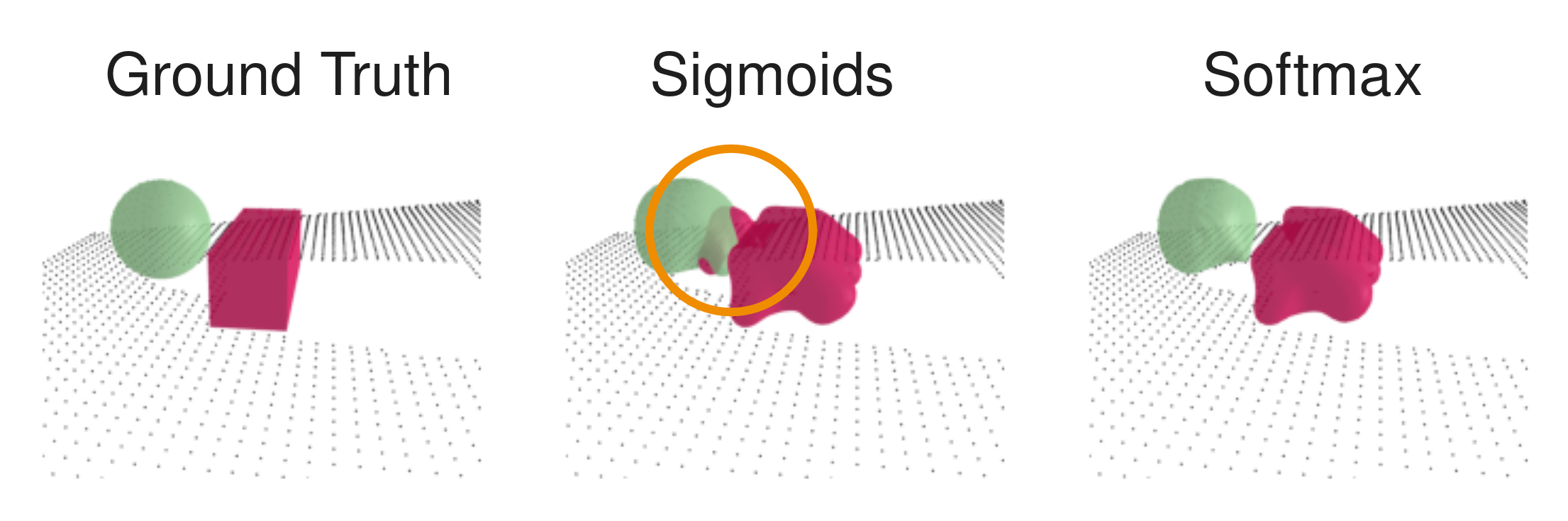}
    \caption{Running a separate sigmoid model per object can cause unwanted intersections between the reconstructions (circled). Our multi-class formulation uses a softmax model that avoids this problem.}
    \label{fig:sigmoidfail}
\end{figure}

Instead of predicting an occupancy map for each object, we phrase our problem as a multi-class mapping problem. This ensures that each point in space can only be occupied by a single object. Without this constraint, reconstructions of objects can intersect each other as shown in \Cref{fig:sigmoidfail}. Formally, we receive observations $\{(\bb x_i, y_i)\}_{i \in [n]}$ where $\bb x_i \in \mathbb R^d$ corresponds to an observed point with segmented class $y_i \in [c]$. We assume $y_i = 1$ denotes $\bb x_i$ being segmented to no specific object and is part of the background or table. We also assume that these observations came from a camera with a known location $\bb o \in \mathbb R^d$. The goal is to build a map function $m: \mathbb R^d \rightarrow \mathbb [0, 1]^{c}$ such that $m(\bb x)$ corresponds to the probability distribution over classes that the point $\bb x$ could belong to.

We would like our map to satisfy that $m(\bb x_i) \approx \bb e_{y_i}$ for all $i$, where $\bb e_{y_i}$ is the one hot encoding of $y_i$. We can infer that for any $\bb x_i$, because the camera ray started at $\bb o$ and terminated at $\bb x_i$, all points in between are unoccupied. We would like our map to reflect this realization. This forms the basis for the negative sampling performed in \cite{senanayake2017bayesian}. We will also assume that objects in the scene are resting on or above a planar surface. While this generally means a table, our method is agnostic to the type of surface. 


\section{Method Overview}\label{sec:overview}


\begin{figure*}[t]
\centering
\includegraphics[width=0.98\linewidth]{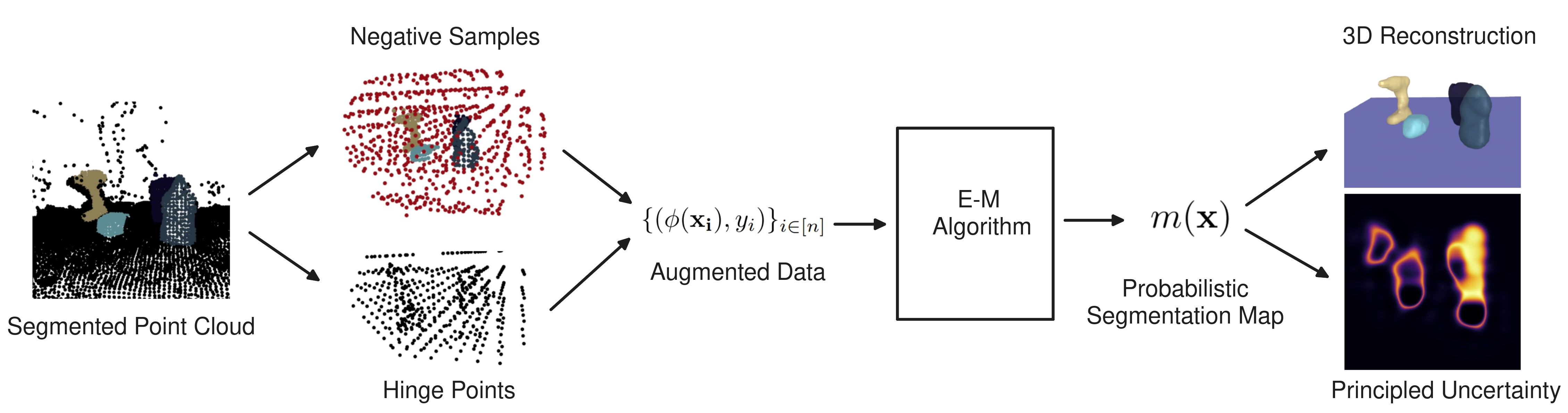}
\caption{Overview of our method, \ourmethod{}. We take a segmented point cloud and output a probabilistic segmentation map over 3D space that can be used for both object reconstruction and principled uncertainty. Our method first generates negative samples and hinge points, then uses these to create an augmented dataset. Then the probabilistic map is constructed by running an EM algorithm over this dataset.}
\label{fig:overview}
\end{figure*}

Our method builds a map $m(\bb x)$ from segmented camera depth observations of a multi-object scene through two main steps. A high level overview is displayed in \Cref{fig:overview}. First, negative sampling is performed as described in \Cref{sec:sampling}, where additional points are added to the observed ones in order to form a new labelled point cloud. During this step, the RANSAC \cite{fischler1981random} algorithm is run in order to recover the surface plane the objects are resting on. The points are also subsampled in order to increase efficiency. We then generate a set of hinge points that are used to construct a feature transform according to \Cref{eq:hinge}. This transform, combined with our sampled points creates a set of augmented data. 

Once we have our tranformed data, we perform Bayesian multi-class regression over the data with an expectation maximization (EM) algorithm. The specific technique makes use of mathematical ideas from \cite{bouchard2007efficient}. The full EM algorithm and model are explored in \Cref{sec:softmax}. Efficiently evaluating $m(\bb x)$ for query $\bb x$ values is also covered in \Cref{sec:softmax}, where we make use of an approximation proposed in \cite{daunizeau2017semi}. The segmentation map produced maps each point in 3D space to a distribution over $c$ classes, where one class denotes not belonging to an object and the other $c-1$ classes denote the segmented objects observed.

Once we have our map, we can use it to evaluate how likely different points are to be in occupied by different objects. This is useful in many motion planning algorithms in order to minimize unwanted collisions. We can also reconstruct the meshes of each object by running the marching cubes algorithm \cite{lorensen1998marching}. These meshes can be used to create a signed distance function, simulate physics, or to visualize the scene. Our map also encodes principled uncertainty about the geometry of the scene which can be used for active inference.


\section{Softmax EM Algorithm}\label{sec:softmax}

\subsection{Training}

To create a Bayesian multi-class map, we consider using a weight matrix $\bb W \in \mathbb R^{c \times m}$ where each row is normally distributed, giving the following likelihood function:
\begin{equation*}
P(y = k | \bb W, \bb x) = \text{softmax}(\bb W \phi(\bb x))_k,
\end{equation*}
where the softmax function is defined as
\begin{equation*}
\text{softmax}(\bb W \phi(\bb x))_k= \frac{\exp(\bb W \phi(\bb x)_k)}{\sum_{i=1}^c \exp(\bb W \phi(\bb x)_i)}.
\end{equation*}

Because a conjugate prior for the softmax likelihood doesn't exist, we must use variational inference to find a posterior Gaussian distribution. In our case, we will maximize a lower bound on the likelihood. A useful inequality for this is given in \cite{bouchard2007efficient}, and is stated in the following theorem:

\begin{theorem}\label{theorem:bound}
From \cite{bouchard2007efficient}. Let $\bb z \in \mathbb R^{c}$, $\alpha \in \mathbb R$, and $\xi \in \mathbb R_+^{c}$. Then the following inequality holds:
\begin{align*}
    \ln \sum_{k=0}^c \exp(\bb z_k) &\leq\alpha + \sum_{k=0}^c \frac{\bb z_k - \alpha - \xi_k}{2} \\
    +& \lambda(\xi_k)((\bb z_k - \alpha)^2 - \xi_k^2) + \ln(1 + \exp(\xi_k)),
\end{align*}
where $\lambda(\xi_k) = ((1 + \exp(-\xi_k))^{-1} - (1 / 2))/2 \xi_k$. 
\end{theorem}

Applying \Cref{theorem:bound} to $\bb z = \bb W \phi(\bb x)$, we can bound the likelihood by introducing the two variational parameters $\alpha$ and $\xi$ with the inequality, 
$$ \ln P(y = k | \bb W, \bb x) \geq \ln Q(y = k | \bb W, \bb x; \alpha, \xi). $$
We can maximize this lower bound and use it as an approximation to the true likelihood by solving the following:
\begin{equation*}    
\text{arg}\max_{\alpha, \xi} \mathbb E_{\bb W} \left[ \ln Q(y = y_i | \bb x_i, \bb W ; \alpha, \xi)\right].
\end{equation*}
This can be analytically solved for $\bb W_k \sim \mathcal N(\mu_k, \Sigma_k)$, yielding the following optimal values found in \cite{bouchard2007efficient}:
\begin{equation}\label{eq:alpha1}
\alpha_i = \frac{\frac{1}{2} (\frac{c}{2} - 1) + \sum_{k=1}^c \lambda(\xi_k) \mu_k^\top \phi(\bb x_i)}{\sum_{k=1}^c \lambda(\xi_k)},
\end{equation}
\begin{equation} \label{eq:xi1}
\xi^2_{i, k} = \phi(\bb x_i)^\top \Sigma_k \phi(\bb x_i) + (\mu^\top_k \phi(\bb x_i))^2 + \alpha_i^2 - 2\alpha_i \mu^\top_k \phi(\bb x_i).
\end{equation}

Due to the inequality used, $P(y=k | \bb W, \bb x; \alpha, \xi)$ is normally distributed for any $\alpha, \xi$ and will be conjugate to our prior weight distribution. Thus, we can compute the closed-form approximate posterior mean $\hat{\mu}$ and covariance $\hat{\Sigma}$ from multiplying $P(y = k | \bb W, \bb x; \alpha, \xi) P(\bb W)$ where $P(\bb W_k) = \mathcal N(\bar{\mu}_k, \bar{\Sigma}_k)$. The update equations mirror those found in \cite{bouchard2007efficient} and are as follows:

\begin{equation}\label{eq:sig1}
\hat{\Sigma}^{-1}_k = \bar{\Sigma}^{-1} + 2 \sum_{i = 1}^n \lambda(\xi_{i,k}) \phi(\bb x_i) \phi(\bb x_i)^\top
\end{equation}

\begin{equation}\label{eq:mu1}
\hat{\mu}_k = \hat{\Sigma}_k \left[ \bar{\Sigma}^{-1}_k \bar{\mu}_k + \sum_{i = 1}^n \left( y_{i, k} - \frac{1}{2} + 2 \alpha_i \lambda(\xi_{i, k}) \right) \phi(\bb x_i) \right].
\end{equation}

\begin{algorithm}[t]
\textbf{Input:} \\
\text{Observed, segmented points $o = \{(\bb x_i, y_i)\}_{i \in [n']}$}
\text{Prior means $\{\bar{\mu}_k\}_{k \in[c]}$ and covariances $\{\bar{\Sigma_k}\}_{k \in[c]}$}
\begin{algorithmic}[1]

\State $\mathcal D \gets \textsc{NegativeSample}(o)$
\State $\phi \gets \textsc{HingePointTransform}(o)$
\State $\xi_{i, k} \gets 1 \text{ for } i \in [m], k \in [c]$
\State $\alpha_i \gets 0 \text{ for } i \in [m]$

\For{$p$ iterations}
\State{$\hat{\Sigma}^{-1} \gets \bar{\Sigma}^{-1} + 2 \sum_i |\lambda(\xi_i)| \phi(\bb x_i) \phi(\bb x_i)^\top$}
\State{$\hat{\mu}_k \gets \hat{\Sigma} \left( \bar{\Sigma}^{-1} \bar{\mu} + \sum_i (y_i - \frac{1}{2} + 2 \alpha_i \lambda (\xi_{i, k})) \phi(\bb x_i) \right)$}
\State{$\alpha_i \gets \textsc{UpdateAlpha}(\xi_i, \bb x_i, \hat{\mu}, \hat{\Sigma})$ with \Cref{eq:alpha1}}
\State{$\xi_{i, k} \gets \textsc{UpdateXi}(\alpha_i, \bb x_i, \hat{\mu}, \hat{\Sigma})$ with \Cref{eq:xi1}}
\EndFor \\
\Return $\hat{\mu}, \hat{\Sigma}$
\end{algorithmic}
\caption{\ourmethod{}}
\label{algo:softmaxbhm}
\end{algorithm}

We can use \Cref{eq:alpha1}, \Cref{eq:xi1}, \Cref{eq:sig1}, and \Cref{eq:mu1} to create an EM algorithm to iterate between calculating our posterior distribution and optimizing our variational parameters. This is shown in \Cref{algo:softmaxbhm}.

\subsection{Inference}

In order to make predictions about new points we need to evaluate the following expectation:
\begin{equation}\label{eq:softmaxexpectation}
\hat{P}(y = k | \bb x) = \mathbb E_{\bb W} \left[ \text{softmax} ({\bb W} \phi(\bb x)) \right]_k.
\end{equation}
There is not a closed form solution to this expectation, so we must approximate it. While we could use sampling to estimate the expectation, we instead use a more computationally efficient approximation. 

As described in \cite{daunizeau2017semi}, we can write the softmax in terms of the sum of sigmoidal terms with the following equality:
\begin{equation*}
\text{softmax}(\bb a)_k = \frac{1}{2 - c + \sum_{i \neq k} \sigma (\bb a_k - \bb a_i)^{-1}},
\end{equation*}
where $c$ is the number of classes. This is then used as motivation for the approximating the expectation with
\begin{equation*}
\mathbb E_{\bb W} \left[ \text{softmax} ({\bb W} \phi(\bb x)) \right]_k \approx \frac{1}{2 - c + \sum_{i \neq k} \mathbb E [\sigma (\Tilde{\bb z}_i)]^{-1}},
\end{equation*}
with $\Tilde{\bb z}_i = [\bb W \phi(\bb x)]_k - \bb [\bb W \phi (\bb x)]_i$. When combined with the sigmoidal approximation in \Cref{eq:sigexpectation}, this becomes an easily computable approximation to \Cref{eq:softmaxexpectation}. 

\section{Object-Centric Negative Sampling}\label{sec:sampling}

Similar to many mapping methods, \ourmethod{} requires negatively sampling points along depth camera rays. The traditional negative sampling used, mentioned in \Cref{subsec:bhm}, is meant for mapping environments where the robot is in an enclosed space and each camera ray is detecting a wall or sufficiently large object. This sampling performs poorly when the goal is to map a relatively small object resting on a tabletop or other surface. To fully utilize the tabletop structure within the environment, we propose a new negative sampling method designed for object-centric mapping. Our sampling method rests on two main realizations:

\begin{figure}[b]
 \centering
    \includegraphics[width=0.95\linewidth]{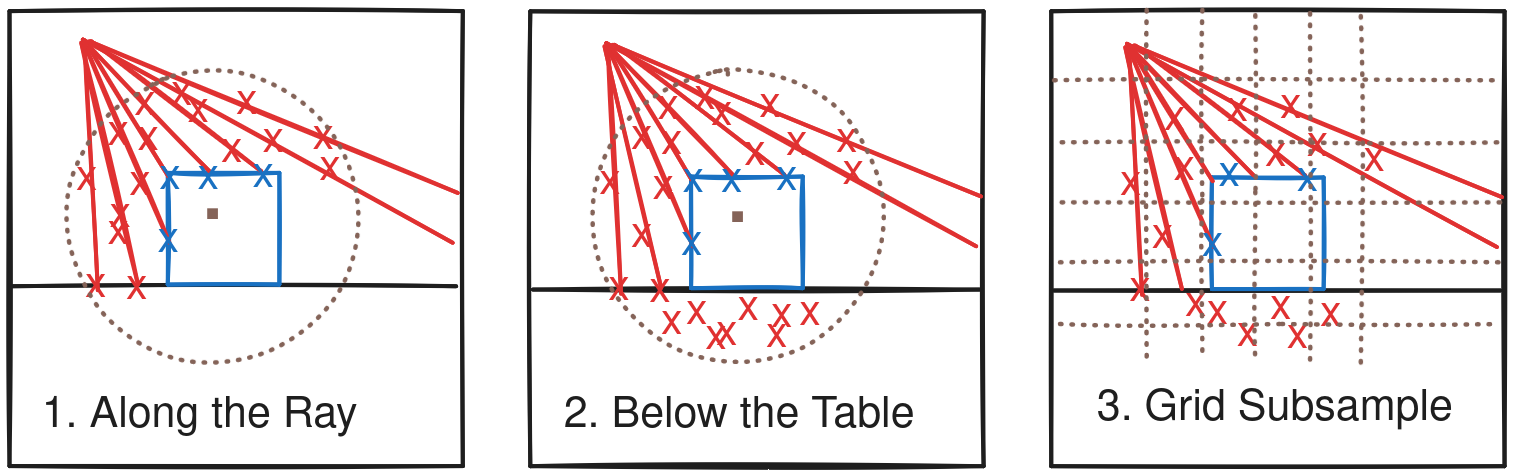}

    \caption{Overview of our sampling method. 1. We perform stratified sampling along camera rays within $r_\text{obj}$ of the object. 2. Points are sampled below the table within $r_\text{obj}$ of the object. 3. Grid subsampling is performed.}
    \label{fig:img2}
\end{figure}

\begin{enumerate}
    \item Along the ray, negative samples are most useful when near known objects.
    \item Points below a surface plane cannot be occupied by objects resting entirely on or above that surface.
\end{enumerate}

We assume we have a segmented point cloud of the scene $\{(\bb x_i, y_i)\}_{i \in [n']}$ where each $y_i$ corresponds to the segmentation label of the respective $\bb x_i$. We also assume a known position of the camera $\bb o$. Our sampling method begins by finding the center of the smallest axis-aligned bounding box that contains all of the segmented points for each individual object in the scene. We denote these centers with $\bb o_k$. We then perform stratified uniform sampling along each ray, only keeping points that are within $r_\text{obj}$ distance from at least one $\bb o_k$. Sampled points within the desired radius of a center are labeled as unoccupied and added to the collection of points for the algorithm.

Next, we run RANSAC \cite{fischler1981random} on the observed point cloud to recover the table plane. Once we have the plane, we uniformly randomly sample points within $r_\text{obj}$ from each object center and keep any such points that fall below the plane. These points are labelled as unoccupied and added to our collection.

Finally, we perform grid subsampling as described in \cite{thomas2019learning} with each label in parallel in order to reduce the number of points our algorithm is fed. In practice, we choose different resolutions to subsample empty points and points on object surfaces. This can dramatically increase the efficiency of our method by removing redundant points. The entire negative sampling process is shown in \Cref{fig:img2}. 

The resulting points are then transformed to construct our set of augmented data. The transform used is induced by a set of hinge points according to \Cref{eq:hinge}. In practice, we choose a set of hinge points consisting of a fixed grid around the scene as well as a fixed number of random points sampled from the surface points of each object.

\section{Experiments}\label{sec:experiments}
\begin{table*}[t]
    \begin{center}
    \begin{tabular}{| c | c | c | c  c | c  c | c  c |}
        \hline
         & & Principled & \multicolumn{2}{c|}{ShapeNet Scenes} & \multicolumn{2}{c|}{YCB Scenes} & \multicolumn{2}{c|}{Objaverse Scenes} \\
        Method & Continuous & Uncertainty & IoU $\uparrow$ & Chamfer (m) $\downarrow$ & IoU $\uparrow$ & Chamfer (m) $\downarrow$ & IoU $\uparrow$ & Chamfer (m) $\downarrow$ \\
        \hline
        Voxel & N & N & 0.198 & 0.014 & 0.324 & 0.018 & 0.336 & 0.024 \\
        PointSDF \cite{van2020learning} & Y & N &  {\bf 0.360} & {\bf 0.010} & 0.460 & 0.015 & 0.347 & 0.025 \\
        \ourmethod{} (ours) & Y & Y & 0.309 & 0.011 & {\bf 0.500} & {\bf 0.012} & {\bf 0.464} & {\bf 0.018} \\
        \hline
    \end{tabular}
    \end{center}
    
    \caption{Quantitative experiments comparing our method to two baseline methods on procedurally generated scenes from benchmark datasets.}
    \label{tab:baseline}
\end{table*}

\begin{table*}[t]
\begin{center}
    \begin{tabular}{| c | c  c | c  c | c  c |}
        \hline
          & \multicolumn{2}{c|}{ShapeNet Scenes} & \multicolumn{2}{c|}{YCB Scenes} & \multicolumn{2}{c|}{Objaverse Scenes} \\
        Method & IoU $\uparrow$ & Chamfer (m) $\downarrow$ & IoU $\uparrow$ & Chamfer (m) $\downarrow$ & IoU $\uparrow$ & Chamfer (m) $\downarrow$ \\
        \hline
        \ourmethod{} w/ BHM Sampling & 0.156 & 0.031 & 0.313 & 0.030 & 0.326 & 0.035 \\
        \ourmethod{} (ours) & {\bf 0.309} & {\bf 0.011} & {\bf 0.500} & {\bf 0.012} & {\bf 0.464} & {\bf 0.018} \\
        \ourmethod{} w/o Under the Table & 0.291 & 0.019 & {\bf 0.500} & 0.014 & 0.439 & 0.024 \\
        \ourmethod{} w/o Stratified Sampling  & 0.145 & 0.024 & 0.294 & 0.023 & 0.291 & 0.029 \\
        \hline
    
    \end{tabular}
\end{center}
\caption{Ablation experiments on our negative sampling method.}
\label{table:samplingablation}
\end{table*}

We perform experiments aimed to answer the following questions: 1. Does our method result in accurate reconstructions? 2. Does our sampling method improve map quality for object-centric mapping? 3. Is our method robust to unknown, noisy scenes? 4. Does our map accurately capture uncertainty about the scene geometry? We test 1 and 2 in \Cref{sec:sim}, 3 in \Cref{sec:qual}, and 4 in \Cref{sec:uncertatin}. We implement \ourmethod{} in PyTorch and run our algorithm on an NVIDIA GeForce RTX 2070 GPU.
\subsection{Baselines and Metrics}
\textbf{Baselines:} We compare our method to two different baselines. The first is a voxel-based heuristic that labels observed unoccupied voxels as unoccupied, observed occupied voxels as their corresponding segmentation id, and unobserved voxels with the same label as the nearest observed voxel. To prevent incorrect predictions below the table plane, we also run RANSAC during our baseline and label all voxels under the plane as unoccupied. We refer to this approach as the {\bf Voxel} baseline. The second baseline is a learning-based approach using a state of the art neural network architecture for continuous object reconstructions in robotics. We take the PointSDF architecture from \cite{van2020learning} and replace the final activation with a sigmoid function to predict occupancy probabilities. We train this model on a dataset of scenes similar to those discussed in \Cref{sec:sim}. The scenes are composed of a subset of the ShapeNet \cite{chang2015shapenet} dataset. Training it on these scenes instead of the original dataset PointSDF was trained on allows it to better function under occlusion and different scales. We refer to this baseline as {\bf PointSDF}.

\textbf{Metrics:} We use two main metrics for comparison: {\bf intersection over union (IoU)} and {\bf Chamfer distance}. IoU is calculated by evaluating points in a fixed grid around each object. Chamfer distance is calculated by first reconstructing the predicted mesh by running the marching cubes algorithm \cite{lorensen1998marching} on a level set of $\hat P(y=1|x) = \tau$ for a chosen $\tau$ of the prediction function. Then, points are sampled from both the predicted mesh and ground truth mesh and the Chamfer distance is calculated between these two point clouds. 

\subsection{Generated Scenes from Benchmark Object Datasets}\label{sec:sim}

In this section, we evaluate our method against the two baseline methods on procedurally generated scenes, from large object datasets. We generate a scene by randomly picking a mesh and placing it at a random pose within predefined bounds with a random scale. We draw meshes from the ShapeNet \cite{chang2015shapenet}, YCB \cite{calli2015ycb}, and Objaverse \cite{deitke2023objaverse} datasets. We generate 100 scenes for each dataset with up to 10 objects in each scene. Objects are placed relatively close together in order to ensure significant occlusion in the scenes. Once the poses have been selected, we simulate physics for a fixed period of time to ensure objects can come to rest. 

Our first experiment on simulated scenes compares our method with the two baselines. Similar to \cite{mescheder2019occupancy}, we use a level set other than $\tau = 0.5$ for constructing the mesh with the neural network. We found $\tau = 0.3$ to provide the best reconstructions for our version of PointSDF. For other methods, we use $\tau = 0.5$. We report the IoU and Chamfer distance in \Cref{tab:baseline}. PointSDF outperforms other methods on the ShapeNet scenes, where the meshes are drawn from the same mesh dataset that it was trained on. On other datasets, our method outperforms PointSDF. This aligns with other work demonstrating that neural networks perform worse the further from the training distribution you get. Because our method has no reliance on a training distribution, it shows consistency across all datasets. Both our method and PointSDF consistently outperform the voxel baseline on most datasets and metrics. The only exception is Chamfer distance on Objaverse scenes, where the voxel baseline outperforms PointSDF. The performance of our method relative to our baselines indicate that our method results in accurate reconstructions

Our second experiment on simulated scenes ablates our negative sampling method. We observe the effect of removing sampling under the table plane and removing the stratified sampling along the ray. In order to remove the stratified sampling, we replace it with taking discrete, fixed steps along each ray instead. We also compare against the original BHM sampling method explained in \cite{senanayake2017bayesian}, where there negative samples are drawn randomly along the whole ray instead of near objects. This is labeled as {\bf BHM Sampling}. The IoU and Chamfer distance are reported in \Cref{table:samplingablation}. Our negative sampling method outperforms alternatives on each dataset and metric. This implies that our proposed sampling method does improve reconstruction quality compared to the others.

The hyperparameters used for the simulated experiments are shown in \Cref{table:hyperparams}. These were kept constant across all procedurally generated datasets and corresponding experiments. 

\begin{figure}[t]
    \centering
    \includegraphics[width=\linewidth]{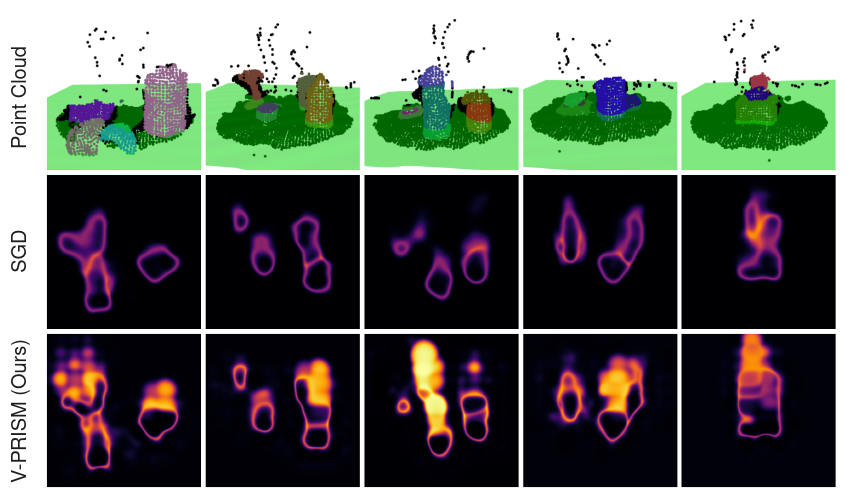}
    \caption{Qualitative comparison of uncertainty. \textbf{Top row:} the observed point cloud with a green plane corresponding to the 2D slice where the heat maps were calculated. We compare a non-probabilistic variant of \ourmethod{} trained with gradient descent (\textbf{middle row}) and our method (\textbf{bottom row}). In the heat maps, the bottom is closer to the camera and the top is farther from the camera. Lighter areas correspond to more uncertainty. Our method predicts high uncertainty in occluded areas of the scene.}
    \label{fig:uncertainty}
\end{figure}

\begin{table}[t!]
    \begin{center}
    
    \begin{tabular}{c|c|c|c}
    \toprule
        Hyperparameters & Value  & Hyperparameters & Value\\
        (Learning) & & (Sampling)& (cm) \\
        \midrule
        kernel type & Gaussian &        grid length & 5.0 \\
        kernel $\gamma$ & 1000 &        sampling $r_\text{obj}$ & 25.0  \\
        surface hinge pts. & 32 &        subsample res. (objects) & 1.0 \\
        iterations & 3 &        subsample res. (empty) & 1.5 \\

    \bottomrule
    \end{tabular}
    \end{center}
    \caption{Hyperparameters for experiments on procedurally generated scenes.}
    \label{table:hyperparams}
\end{table}

\subsection{Real World Scenes}\label{sec:qual}

\begin{figure*}[t]
    \centering
    \includegraphics[width=0.96\linewidth]{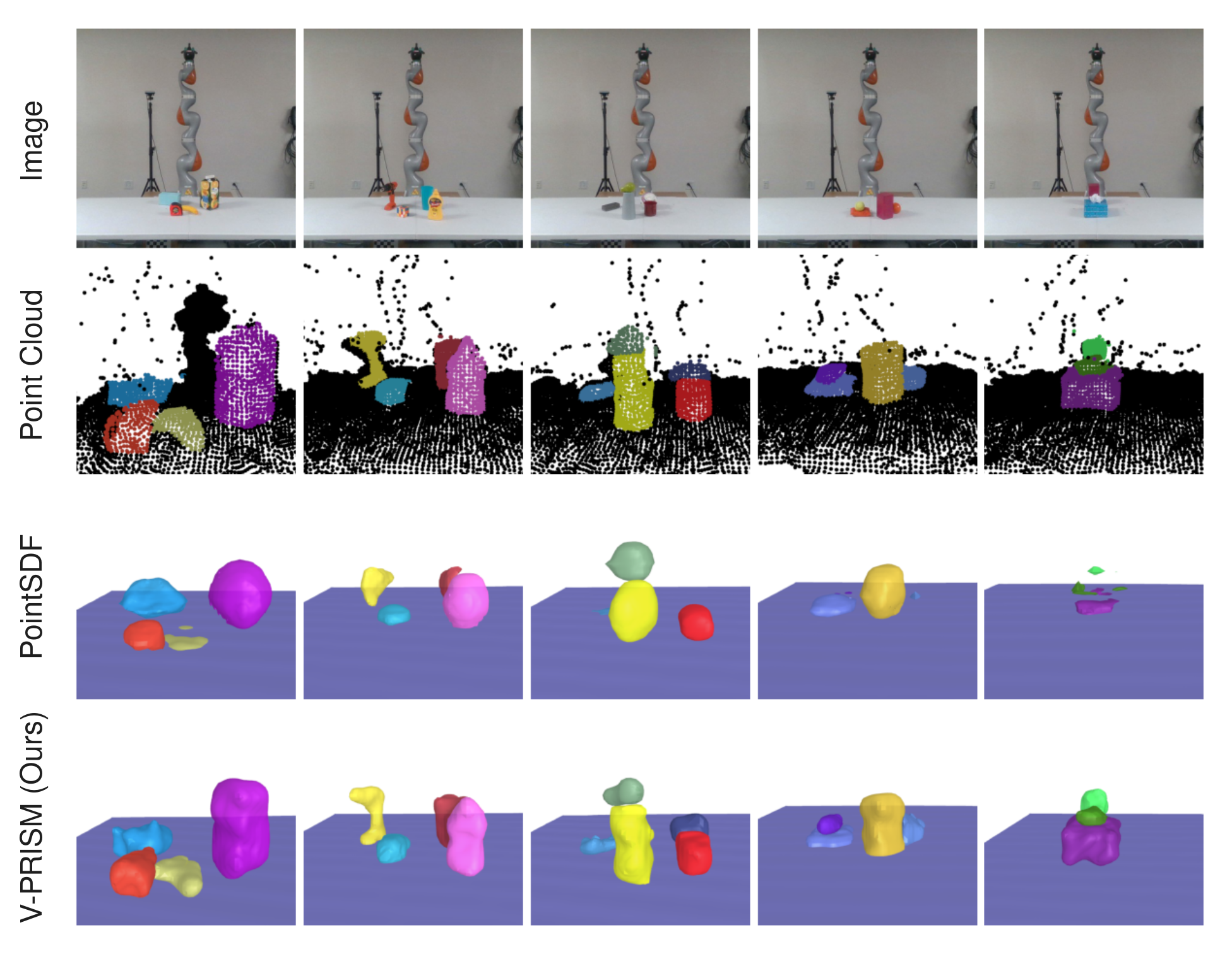}
    \caption{Qualitative comparisons with PointSDF reconstructions. \textbf{First row:} RGB images. \textbf{Second row:} the segmented point cloud used as input. \textbf{Third row:} PointSDF reconstructions. \textbf{Last row:} \ourmethod{}'s (our method) reconstructions. \ourmethod{} results in quality reconstructions on noisy scenes.}
    \label{fig:qual_pointsdf}
\end{figure*}

We evaluate our method by qualitatively comparing reconstructions on real world scenes. We use a Intel RealSense D415C camera to obtain point clouds of tabletop scenes. In order to get accurate segmentations of the scene, we use the Segment Anything Model (SAM) \cite{kirillov2023segment}. We compute on five scenes consisting of multiple objects. We compare our method to PointSDF. The qualitative comparison can be seen in \Cref{fig:qual_pointsdf}. Because these scenes are significantly more noisy than simulated scenes, PointSDF struggles to coherently reconstruct the scene. In contrast, our method is capable of producing quality reconstructions even with very noisy input point clouds. This suggests that our method is capable of bridging the sim to real gap and is robust to unknown, noisy scenes. 

\subsection{Principled Uncertainty}\label{sec:uncertatin}

To show how our model captures uncertainty about the scene, we need a way to quantify uncertainty. We use the entropy of our map at each point in space as a measurement of uncertainty:
\begin{equation*}
    H_m(\bb x) = - \sum_{k=1}^c \hat P(y = c | \bb x) \ln \hat P(y = c | \bb x).
\end{equation*}
This is maximized when the model predicts a uniform distribution over classes and minimized when the model predicts a single class with a probability of 1.

We compare our method with an alternate non-Bayesian version of our method, where we train a single weight vector with stochastic gradient descent (SGD) instead of the EM algorithm, to minimize the negative log-likelihood of our augmented data.

To visualize this uncertainty, we calculate this uncertainty over a 2D slice from each of our 5 real world scenes. The heat maps for each slice can be seen in \Cref{fig:uncertainty}. Qualitatively, we can see that our method obtains high uncertainty values in occluded sections of the scene. This contrasts to the non-probabilistic model that does not accurately capture uncertainty about occluded regions. The heat maps showing occlusion-aware uncertainty suggest our model captures principled and accurate uncertainty measures.

\section{Conclusion and Future Work}\label{sec:conclusion}  


Principled uncertainty is necessary for the safety of many robotics tasks. We proposed a framework for robustly constructing multi-class 3D maps of tabletop scenes named \ourmethod{}. Our method works by iterating an EM algorithm on augmented data to produce a volumetric Bayesian segmentation map. To fully incorporate the information from received depth measurements of a tabletop scene, we proposed a novel negative sampling technique. The resulting map was shown to have desirable properties including quality reconstructions and accurate uncertainty measures through both quantitative experiments in simulation and qualitative experiments with real-world, noisy scenes. Future directions of this work include: (1) using our method's uncertainty to inform active learning; (2) extending V-PRISM to represent dynamic tabletop scenes.

\bibliographystyle{ieeetr}
\bibliography{refs} 

\end{document}